\pdfoutput=1
\documentclass[oribibl]{llncs}

\usepackage{booktabs}
\usepackage{framed}
\usepackage{dirtytalk}
\usepackage{soul}

\renewcommand\hl[1]{#1} 

\usepackage[utf8]{inputenc} 
\usepackage[T1]{fontenc}    
\usepackage{hyperref}       
\usepackage{url}            
\usepackage{booktabs}       
\usepackage{amsfonts}       
\usepackage{nicefrac}       
\usepackage{microtype}      
\usepackage{xcolor}         
\usepackage[pdftex]{graphicx}
\usepackage{subcaption}
\usepackage{float}
\usepackage{tabularx} 

\title{
Visconde: Multi-document QA with GPT-3 and Neural Reranking
}

\author{Jayr Pereira\inst{1,2}\orcidID{0000-0001-5478-438X} \and
Robson Fidalgo\inst{2}\orcidID{0000-0002-4714-2933} \and
Roberto Lotufo\inst{1}\orcidID{0000-0002-5652-0852}
\and
Rodrigo Nogueira\inst{1}\orcidID{0000-0002-2600-6035}}

\authorrunning{Pereira et al.}
\institute{NeuralMind, Brazil\\
\email{\{jayr.pereira,roberto,rodrigo.nogueira\}@neuralmind.ai}\\
\and
Centro de Informática, Universidade Federal de Pernambuco, Brazil\\
\email{\{jap2,rdnf\}@cin.ufpe.br}}

\begin{document}

\maketitle

\begin{abstract}

This paper proposes a question-answering system that can answer questions whose supporting evidence is spread over multiple (potentially long) documents. The system, called Visconde, uses a three-step pipeline to perform the task: decompose, retrieve, and aggregate. The first step decomposes the question into simpler questions using a few-shot large language model (LLM). Then, a state-of-the-art search engine is used to retrieve candidate passages from a large collection for each decomposed question. In the final step, we use the LLM in a few-shot setting to aggregate the contents of the passages into the final answer. The system is evaluated on three datasets: IIRC, Qasper, and StrategyQA. Results suggest that current retrievers are the main bottleneck and that readers are already performing at the human level as long as relevant passages are provided. The system is also shown to be more effective when the model is induced to give explanations before answering a question. Code is available at \url{https://github.com/neuralmind-ai/visconde}.

\end{abstract}

\section{Introduction} \label{sec:introduction}

In recent years, question-answering (QA) tasks that use relatively short contexts (e.g., a paragraph) have seen remarkable progress in multiple domains~\cite{khashabi2020unifiedqa,khashabi2022unifiedqa}. However, in many cases, the necessary information to answer a question is spread over multiple documents or long ones~\cite{yang2018hotpotqa,perez2020unsupervised,ferguson-etal-2020-iirc,geva2021strategyqa}.
To solve this task, QA models are based on a pipeline comprised of a retriever and a reader component~\cite{karpukhin-etal-2020-dense,ferguson-etal-2022-retrieval,trivedi2022teching,sachan2022questions}. Most of these approaches rely on fine-tuning large language models on supervised datasets, which may be available for a variaty of domains. Other approaches use Transformers for long sequences like LongT5 \cite{guo2021longt5} to process the context document and the question at once \cite{ray2022mod,xiong2022longt5}, which might not scale to longer sequences (e.g., documents with hundreds of pages).

The few-shot capability of LLMs may reduce the costs for solving QA tasks, as it allows one to implement QA systems for different domains without needing a specific annotated dataset. 
In addition, recent studies showed that adding a chain-of-thought (CoT) reasoning step before answering significantly improves LLMs' zero or few-shot effectiveness on diverse QA benchmarks \cite{kojima2022llm}.
\hl{In this work, we propose Visconde,\footnote{The name is a homage to \textit{Visconde de Sabugosa} a fictional character invented by Monteiro Lobato that is a corn cob doll whose wisdom comes from reading books.} a QA system that combines a state-of-the-art retriever and a few-shot (CoT) approach to induce an LLM to generate the answer as a generative reader.
The retriever is a multi-stage pipeline that uses BM25~\cite{Robertson1994OkapiAT} to select candidate documents followed by a monoT5 reranker~\cite{nogueira2020document}. The reader uses GPT-3 \cite{openai2020gpt3} in a few-shot setting that reason over the retrieved records to produce an answer.
We induce CoT by asking the model to explain how the evidence documents can answer the question. 
Our system rivals state-of-the-art supervised models in three datasets: IIRC, Qasper, and StrategyQA.}

Our main contribution is to show that current multi-document QA systems are close to human-level performance \textit{as long as ground truth contexts are provided as input to the reader}. When a SOTA retriever selects the context, we observe a significant drop in effectiveness. Thus, we argue future work on multi-document QA should focus on improving retrievers.

\section{Related Work}

Most approaches for multi-document QA are typically based on a retriever followed by a reader component \cite{zhu2021rr}. The retriever aims to select relevant documents for given a question, while the reader seeks to infer the final answer from them.
Recent studies used dense retrievers \cite{trivedi2022teching,ferguson-etal-2020-iirc} or commercial search engines \cite{lazaridou2022internet} for this task. For the reader component, some studies used sequence-to-sequence models to generate natural language answers~\cite{lewis2020retrieval,xiong2020answering,izacard-grave-2021-leveraging}, numerical reasoning models adapted to reason also over text~\cite{ferguson-etal-2022-retrieval,trivedi2022teching,ferguson-etal-2020-iirc}, or LLMs to aggregate information from multiple documents \cite{webgpt,lazaridou2022internet}. 

Recent work enriches this pipeline by adding components to perform query decomposition~\cite{press2022measuring,das2019multistep,feldman-el-yaniv-2019-multi,boosting2022boerschinger,zeroshot2022huebscher} or evidence retrieved by a web search engine~\cite{webgpt,press2022measuring,lazaridou2022internet}. 
Our work is similar to these, but we focus on evaluating the limitations of this method and found that the retrieval component needs more work.

\section{Our Method: Visconde} \label{sec:method}

\begin{figure}[t]
\centering
\includegraphics[width=\textwidth]{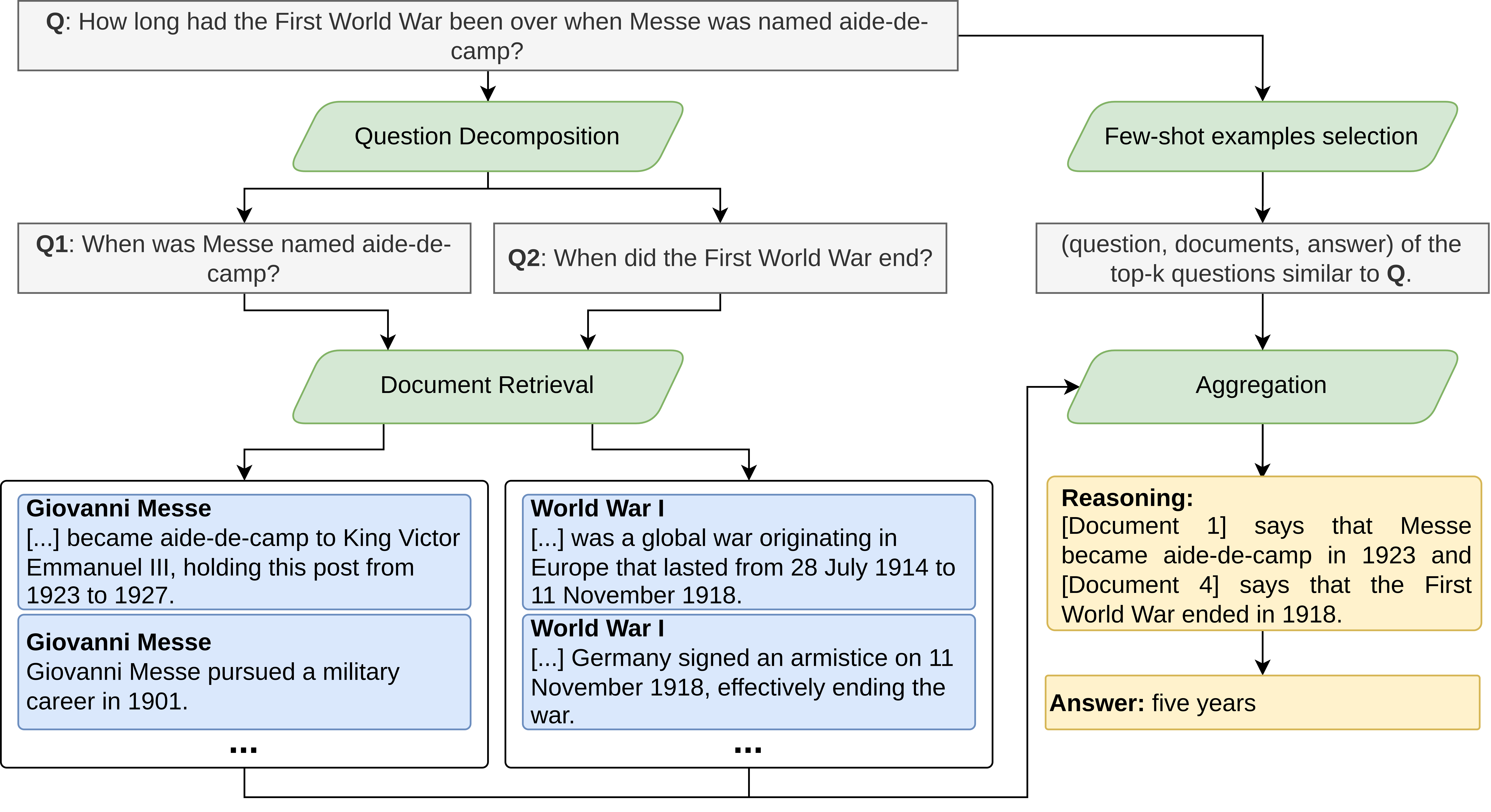}
\caption{Visconde QA flow.}
\label{fig:search_strategy}
\end{figure}

Visconde is a multi-document QA system that has three main steps: 
\textit{Question decomposition}, \textit{Document Retrieval}, and \textit{Aggregation}. As illustrated in Figure \ref{fig:search_strategy}, the system first decomposes the user question when necessary and searches for relevant documents to answer the subquestions. The retrieved documents are the basis for generating an explanation and a final answer using an LLM.


\noindent \textbf{Question Decomposition:} We use GPT-3 (text-davinci-002) with five in-context examples for question decomposition. 
In Figure \ref{fig:search_strategy} we show an example of a question that needs to be decomposed (\textbf{Q}) extracted from the IIRC dataset \cite{ferguson-etal-2020-iirc} and the subquestions generated by the model (\textbf{Q1} and \textbf{Q2}).
\hl{The five examples used as few-shot in the prompt were randomly selected from the training set of StrategyQA dataset \cite{geva2021strategyqa}, which has questions and decomposed subquestions.
}


\noindent \textbf{Document Retrieval:} For document retrieval, 
we used a strategy divided into three main steps: 1) document indexing -- we create an inverted index using Pyserini \cite{Lin2021pyserini}; 2) candidates retrieval -- we use the Pyserini implementation of the BM25 algorithm to retrieve candidate documents. 
3) document reranking -- we rerank the top-1000 documents retrieved using a sequence-to-sequence model designed for reranking, the monoT5 model~\cite{nogueira2020document}, an adaptation of the T5 model \cite{raffel2020exploring}.\footnote{We used the 3 billion parameters version, whose checkpoint is available at \url{https://huggingface.co/castorini/monot5-3b-msmarco-10k}}
\hl{monoT5 receives as input a sequence with the document and the query, and a softmax function is applied only on the logits calculated by T5 to the tokens \texttt{true} and \texttt{false}. The log probability of the token \texttt{true} is used as the document relevance score given the question. The output is a list of documents ranked by the relevance scores.}


\noindent \textbf{Aggregation:} We use GPT-3 (text-davinci-002) as a few-shot learner for the aggregation step.
Different studies have shown that the effectiveness of LLM can be improved by inducing it to first generate reasoning steps before answering a question~\cite{wei2022CoT,wang2022,kojima2022llm,creswell2022faithful}.
We use CoT to induce the LLM to reason over multiple documents and answer a question, as shown in the example in Figure \ref{fig:prompt}.
\hl{In our prompt, each example has a list of context documents (e.g., [Document 1]), a question, an evidence paragraph, and the answer. 
The context documents of the target example are the top-k documents from the retriever step. When the question is decomposed, we use the top-k documents from each subquestion.
For the target example, an evidence paragraph is not provided, and the LLM must generate it, as well as a final answer.}

\begin{figure}[t]
    \caption{Reasoning prompt example. The \textbf{bold} text is the model's output.}
    \label{fig:prompt}
    \begin{framed}
    \fontsize{7pt}{8pt}\selectfont
    \textcolor{gray}{-- Few-shot examples omitted due to space --}
    \newline
    \newline
    Example 5:
    \newline
    \newline
    [Document 1]: Title: Giovanni Messe. Content: [...] became aide-de-camp to King Victor Emmanuel III, holding this post from 1923 to 1927.
    \newline
    \newline
    \textcolor{gray}{-- Documents omitted due to space --}
    \newline
    \newline
    [Document 5]: Title: World War I. Content: [...] Germany signed an armistice on 11 November 1918, effectively ending the war.
    \newline
    \newline
    Question: How long had the First World War been over when Messe was named aide-de-camp?
    \newline
    \newline
    Evidence: \textbf{According to [Document 5], the First World War ended on November 11th, 1918. According to [Document 1], Messe was named aide-de-camp on 1923. Therefore, the First World War had been over for 5 years when Messe was named aide-de-camp.}
    \newline
    \newline
    Answer: \textbf{5 years}
    \end{framed}
\end{figure}{}

We tested two approaches for prompt construction: 1) using static prompts with a pre-defined list of examples; and 2) using dynamic prompts by selecting in-context examples that are similar to the test example. For dynamic prompts, we encode the questions from the training dataset using SentenceTransformers \cite{reimers2019sbert} and apply a KNN algorithm to find the $k$ most similar to the test question, as Liu et al.~\cite{Liu2021inContext} did for other tasks.\footnote{We used this model as our sentence encoder: \textit{sentence-transformers/msmarco-bert-base-dot-v5}.} 

\section{Experiments} \label{sec:experiments}


\subsection{IIRC}

The Incomplete Information Reading Comprehension (IIRC) dataset \cite{ferguson-etal-2020-iirc} consists of information-seeking questions that require retrieving the necessary information missing from the original context. Each original context is a paragraph from the English Wikipedia, which comes with a set of links to other Wikipedia pages. 

\textbf{Pre-processing:} we used the dynamic prompt described in Section \ref{sec:method}. For this, we automatically generated reasoning paragraphs for 10\% of the IIRC training set (1340 questions) using GPT-3. In addition, we processed the context articles provided by the dataset to create a searchable index.

\textbf{Procedure:} following the framework depicted in Figure \ref{fig:search_strategy}, we first decomposed the questions from the IIRC test set. We performed document retrieval on a database of Wikipedia documents provided in the dataset.
In the aggregation step, we applied four methods:
1) using GPT-3 without CoT and providing the links and the ground truth contexts, i.e., skipping the document retrieval step;
2) using the reasoning step with the links and ground truth contexts;
3) using reasoning step over the intersection of retrieved documents and the documents cited by the main context; and
4) reasoning over the documents retrieved from the entire Wikipedia subset provided by the dataset.



\subsection{Qasper}

Qasper \cite{Dasigi2021ADO} is an information-seeking QA dataset over academic research papers. 
The task consists of retrieving the most relevant evidence paragraph for each question and answering the question.


\textbf{Procedure:} we did not apply query decomposition because the questions in this dataset are closed-ended, i.e., they do not require decomposition as they are grounded in a single paper of interest \cite{Dasigi2021ADO}. 
For example, the question \say{How is the text segmented?} only makes sense concerning its grounded paper.
Besides, we skipped the BM25 step for document retrieval as the monoT5 reranker can score each paragraph in the paper in a reasonable time. 
The document retrieval step consists of reranking the paper's paragraphs based on the question and choosing the top five as context documents. 
We did not notice any advantage in using a dynamic prompt in this dataset.

\subsection{StrategyQA}

StrategyQA \cite{geva2021strategyqa} is a dataset focused on open-domain questions that require reasoning steps. 
This dataset has three tasks: 1) question decomposition, measured using a metric called SARI, generally used to evaluate automatic text simplification systems \cite{Xu2016}; 2) evidence paragraph retrieval, measured as the recall of the top ten retrieved results; and 3) question answering, measured in terms of accuracy.

\textbf{Pre-processing:} we did not generate reasoning paragraphs for the training examples since the context comprises long paragraphs that exceed the model input size limit (4000 tokens). We processed the context articles provided by the dataset to create a searchable index, by splitting the articles into windows of three sentences each.

\textbf{Procedure:} we applied question decomposition and performed retrieval using the approach described in Section~\ref{sec:method}. We used the top five retrieved documents for each decomposed question as the context in the reading step. 

\subsection{Results} \label{sec:results}


\begin{table}[h!]
\footnotesize
\centering
\caption{Visconde and similar methods results on IIRC, Qasper and StrategyQA.}
\label{tab:results}
\begin{subtable}[t]{\textwidth}
\footnotesize
\centering
\begin{tabular*}{\textwidth}{@{}l@{\extracolsep{\fill}}cc@{}}
\toprule
\multicolumn{3}{c}{\textbf{IIRC}} \\ \midrule
Model                                                                              & F1   & EM   \\ \midrule
Human                                                                              & 88.4 & 85.7 \\
\midrule
\textit{Finetuned}\\
Ferguson et al.~\cite{ferguson-etal-2020-iirc}               & 31.1 & 27.7 \\
Ferguson et al.~\cite{ferguson-etal-2020-iirc} Linked pages  & 32.5 & 29.0 \\
Ferguson et al.~\cite{ferguson-etal-2020-iirc} Gold Ctx      & 70.3 & 65.6 \\
PReasM (pretrain + finetuning) \cite{trivedi2022teching} Gold Ctx & -    & 73.3 \\
PReasM (pretrain + finetuning) \cite{trivedi2022teching}          & -    & 47.2 \\
Sup$_{A+QA}$ (supervised) \cite{ferguson-etal-2022-retrieval}     & 51.6 & -    \\
\midrule
\textit{Few-shot}\\
Visconde (4-shot dynamic prompt) Gold Ctx and CoT                                  & 84.2 & 74.7 \\
Visconde (4-shot dynamic prompt) Gold Ctx                                          & 80.3 & 70.0 \\
Visconde (4-shot static prompt) Gold Ctx                                           & 74.3 & 62.7 \\
Visconde (4-shot dynamic prompt) Linked pages                                      & 48.2 & 40.7 \\
Visconde (4-shot dynamic prompt) CoT                                               & 47.9 & 40.0 \\ 
\end{tabular*}
\end{subtable}
\begin{subtable}[t]{\textwidth}
\centering
\begin{tabular*}{\textwidth}{@{}l@{\extracolsep{\fill}}cccccc@{}}
\toprule
\multicolumn{7}{c}{\textbf{Qasper}} \\ \midrule
Model                                             & Extractive & Abstractive & Boolean & Unanswerable   & Evidence F1 & Answer F1 \\ \midrule
Human                                             & 58.9     & 39.7      & 79.0  & 69.4 & 71.6        & 60.9      \\
LED-base                                          & 30.0     & 15.0      & 68.9   & 45.0 & 29.9        & 33.6      \\
SOTA sup. \cite{xiong2022longt5} & -          & -           & -       & -      & -           & 53.1      \\
Visconde                  & 52.3      & 21.7       & 86.2   & 48.3  & 38.5        & 49.1      \\ 
\end{tabular*}
\end{subtable}
\begin{subtable}[t]{\textwidth}
\centering

\begin{tabular*}{\textwidth}{@{}l@{\extracolsep{\fill}}ccc@{}}
\toprule
\multicolumn{4}{c}{\textbf{StrategyQA}} \\ \midrule
Model                                                                         & Acc   & Recall@10 & SARI \\ \midrule
Human                                                                         & 87.00 & 0.586       & -    \\
Baseline                                                                      & 63.60 & 0.195       & -    \\
Leaderboard's SOTA                                                            & 69.80 & 0.537       & 0.555  \\
GOPHER-280B OB$^{PoE}_{Google}$ \cite{lazaridou2022internet} & 66.20 & -         & -    \\
Visconde (1-shot, static prompt, gold evidences)                              & 73.80 & -         & -    \\
Visconde (1-shot, static prompt)                                              & 69.43 & 0.331       & 0.570  \\ \bottomrule
\end{tabular*}
\end{subtable}
\end{table}

In Table~\ref{tab:results}, we present the results of our experiments.
First, we show the results obtained in the IIRC dataset.
Our approach outperforms the baselines (i.e., Ferguson et al. \cite{ferguson-etal-2020-iirc}'s) in different settings: 
1) Using the gold context searched by humans (Gold Ctx);
2) Searching for context in the links the dataset provides (Linked pages); and
3) Searching for contexts in the entire dataset.
We report Visconde's results with and without CoT and using a static prompt instead of a dynamic one. The dynamic prompt leads to better performance in this dataset.
When using the gold contexts, Visconde approaches human performance in terms of F1. However, when the system has to search for context, performance decreases. Also, the system performs better when using CoT.
\hl{With CoT, Visconde tends to perform better in questions requiring basic arithmetic operations, which is consistent with the literature \cite{kojima2022llm,wei2022CoT}.
By inspecting the model output in relation to the expected answer, we noticed that in some cases the system answers questions marked as unanswerable by the human annotators. This may occur 1) because the retriever found a relevant document containing the answer or 2) because GPT-3 answers the question even when the necessary information is not in the context. Different ways to write numerals may also affect the results. For example, the model might answer \say{five years}, while \say{5 years} is expected.}




For Qasper, we present the results in terms of F1 for the answer and the evidence. The LED-base model is the baseline \cite{Dasigi2021ADO}. The SOTA model is the model proposed by Xiong et al. \cite{xiong2022longt5}. 
Visconde outperformed the baseline but did not surpass the SOTA model. 
Xiong et al. \cite{xiong2022longt5}'s model is a long Transformer fine-tuned on the task. 
Regarding evidence F1, our system outperforms the baseline, but there is still a gap between our performance and human performance. 
\hl{Visconde had a high performance on the boolean questions but a low score on the abstractive ones. The table shows that even the human result is lower for the abstractive question than other types.}


For StrategyQA, we present the results in terms of answer accuracy, evidence recall@10, and SARI for question decomposition. Automated methods are still far from human performance. 
Our approach outperforms the baselines presented in the paper \cite{geva2021strategyqa} in terms of answer accuracy and evidence recall@10. We also outperform the leaderboard's SOTA model\footnote{\url{https://leaderboard.allenai.org/strategyqa/submissions/public}. Accessed on July 20 2022.} in the quality of the questions decomposition measured with SARI. 
However, we did not surpass SOTA's recall@10 in retrieving the appropriate evidence paragraphs and coming close in the answer accuracy. We also outperform Lazaridou et al.~'s approach \cite{lazaridou2022internet}, which also uses a few-shot LLM.




\section{Conclusion}

This paper describes a system for multi-document question answering that uses a passage reranker to retrieve documents and large language models to reason over them and compose an answer. 
Our system rivals state-of-the-art supervised models in three datasets: IIRC, Qasper, and StrategyQA.
Our results suggest that using GPT-3 as a reader is close to human-level performance as long as relevant passages are provided, while current retrievers are the main bottleneck. We also show that inducing the model to give explanations before answering a question improves effectiveness.

\section*{Aknowledgements}

This research was partially supported by Fundação de Amparo à Pesquisa do Estado de São Paulo (FAPESP) (project id 2022/01640-2) and by Coordenação de Aperfeiçoamento de Pessoal de Nível Superior (CAPES) (Grant code: 88887.481522/2020-00). We also thank Centro Nacional de Processamento de Alto Desempenho (CENAPAD-SP) and Google Cloud for computing credits.

\bibliographystyle{splncs04}

\bibliography{main}

\begin{thebibliography}{10}
\providecommand{\url}[1]{\texttt{#1}}
\providecommand{\urlprefix}{URL }
\providecommand{\doi}[1]{https://doi.org/#1}

\bibitem{boosting2022boerschinger}
Boerschinger, B., Buck, C.C.F., Espeholt, L.J.G., Adolphs, L., Saralegui, L.S.,
  Ciaramita, M., Huebscher, M.C., Sessa, P.G., Rothe, S., Hofmann, T., Kilcher,
  Y.: Boosting search engines with interactive agents. Transactions on Machine
  Learning Research  (2022), \url{https://openreview.net/pdf?id=0ZbPmmB61g}

\bibitem{openai2020gpt3}
Brown, T., Mann, B., Ryder, N., Subbiah, M., Kaplan, J.D., Dhariwal, P.,
  Neelakantan, A., Shyam, P., Sastry, G., Askell, A., Agarwal, S.,
  Herbert-Voss, A., Krueger, G., Henighan, T., Child, R., Ramesh, A., Ziegler,
  D., Wu, J., Winter, C., Hesse, C., Chen, M., Sigler, E., Litwin, M., Gray,
  S., Chess, B., Clark, J., Berner, C., McCandlish, S., Radford, A., Sutskever,
  I., Amodei, D.: Language models are few-shot learners. In: Larochelle, H.,
  Ranzato, M., Hadsell, R., Balcan, M., Lin, H. (eds.) Advances in Neural
  Information Processing Systems. vol.~33, pp. 1877--1901. Curran Associates,
  Inc. (2020),
  \url{https://proceedings.neurips.cc/paper/2020/file/1457c0d6bfcb4967418bfb8ac142f64a-Paper.pdf}

\bibitem{creswell2022faithful}
Creswell, A., Shanahan, M.: Faithful reasoning using large language models.
  arXiv preprint arXiv:2208.14271  (2022)

\bibitem{das2019multistep}
Das, R., Dhuliawala, S., Zaheer, M., McCallum, A.: Multi-step retriever-reader
  interaction for scalable open-domain question answering (2019).
  \doi{10.48550/ARXIV.1905.05733}, \url{https://arxiv.org/abs/1905.05733}

\bibitem{Dasigi2021ADO}
Dasigi, P., Lo, K., Beltagy, I., Cohan, A., Smith, N.A., Gardner, M.: A dataset
  of information-seeking questions and answers anchored in research papers. In:
  Proceedings of the 2021 Conference of the North American Chapter of the
  Association for Computational Linguistics: Human Language Technologies. pp.
  4599--4610. Association for Computational Linguistics, Online (Jun 2021).
  \doi{10.18653/v1/2021.naacl-main.365},
  \url{https://aclanthology.org/2021.naacl-main.365}

\bibitem{feldman-el-yaniv-2019-multi}
Feldman, Y., El-Yaniv, R.: Multi-hop paragraph retrieval for open-domain
  question answering. In: Proceedings of the 57th Annual Meeting of the
  Association for Computational Linguistics. pp. 2296--2309. Association for
  Computational Linguistics, Florence, Italy (Jul 2019).
  \doi{10.18653/v1/P19-1222}, \url{https://aclanthology.org/P19-1222}

\bibitem{ferguson-etal-2020-iirc}
Ferguson, J., Gardner, M., Hajishirzi, H., Khot, T., Dasigi, P.: {IIRC}: A
  dataset of incomplete information reading comprehension questions. In:
  Proceedings of the 2020 Conference on Empirical Methods in Natural Language
  Processing (EMNLP). pp. 1137--1147. Association for Computational
  Linguistics, Online (Nov 2020). \doi{10.18653/v1/2020.emnlp-main.86},
  \url{https://aclanthology.org/2020.emnlp-main.86}

\bibitem{ferguson-etal-2022-retrieval}
Ferguson, J., Hajishirzi, H., Dasigi, P., Khot, T.: Retrieval data augmentation
  informed by downstream question answering performance. In: Proceedings of the
  Fifth Fact Extraction and VERification Workshop (FEVER). pp.~1--5.
  Association for Computational Linguistics, Dublin, Ireland (May 2022).
  \doi{10.18653/v1/2022.fever-1.1},
  \url{https://aclanthology.org/2022.fever-1.1}

\bibitem{geva2021strategyqa}
Geva, M., Khashabi, D., Segal, E., Khot, T., Roth, D., Berant, J.: {Did
  Aristotle Use a Laptop? A Question Answering Benchmark with Implicit
  Reasoning Strategies}. Transactions of the Association for Computational
  Linguistics  \textbf{9},  346--361 (04 2021). \doi{10.1162/tacl\_a\_00370},
  \url{https://doi.org/10.1162/tacl\_a\_00370}

\bibitem{guo2021longt5}
Guo, M., Ainslie, J., Uthus, D., Ontanon, S., Ni, J., Sung, Y.H., Yang, Y.:
  Longt5: Efficient text-to-text transformer for long sequences (2021).
  \doi{10.48550/ARXIV.2112.07916}, \url{https://arxiv.org/abs/2112.07916}

\bibitem{zeroshot2022huebscher}
Huebscher, M.C., Buck, C., Ciaramita, M., Rothe, S.: Zero-shot retrieval with
  search agents and hybrid environments (2022).
  \doi{10.48550/ARXIV.2209.15469}, \url{https://arxiv.org/abs/2209.15469}

\bibitem{izacard-grave-2021-leveraging}
Izacard, G., Grave, E.: Leveraging passage retrieval with generative models for
  open domain question answering. In: Proceedings of the 16th Conference of the
  European Chapter of the Association for Computational Linguistics: Main
  Volume. pp. 874--880. Association for Computational Linguistics, Online (Apr
  2021). \doi{10.18653/v1/2021.eacl-main.74},
  \url{https://aclanthology.org/2021.eacl-main.74}

\bibitem{karpukhin-etal-2020-dense}
Karpukhin, V., Oguz, B., Min, S., Lewis, P., Wu, L., Edunov, S., Chen, D., Yih,
  W.t.: Dense passage retrieval for open-domain question answering. In:
  Proceedings of the 2020 Conference on Empirical Methods in Natural Language
  Processing (EMNLP). pp. 6769--6781. Association for Computational
  Linguistics, Online (Nov 2020). \doi{10.18653/v1/2020.emnlp-main.550},
  \url{https://aclanthology.org/2020.emnlp-main.550}

\bibitem{khashabi2020unifiedqa}
Khashabi, D., Min, S., Khot, T., Sabhwaral, A., Tafjord, O., Clark, P.,
  Hajishirzi, H.: Unifiedqa: Crossing format boundaries with a single qa system
  (2020)

\bibitem{khashabi2022unifiedqa}
Khashabi, D., Kordi, Y., Hajishirzi, H.: Unifiedqa-v2: Stronger generalization
  via broader cross-format training. arXiv preprint arXiv:2202.12359  (2022)

\bibitem{kojima2022llm}
Kojima, T., Gu, S.S., Reid, M., Matsuo, Y., Iwasawa, Y.: Large language models
  are zero-shot reasoners (2022). \doi{10.48550/ARXIV.2205.11916},
  \url{https://arxiv.org/abs/2205.11916}

\bibitem{lazaridou2022internet}
Lazaridou, A., Gribovskaya, E., Stokowiec, W., Grigorev, N.: Internet-augmented
  language models through few-shot prompting for open-domain question
  answering. arXiv preprint arXiv:2203.05115  (2022)

\bibitem{lewis2020retrieval}
Lewis, P., Perez, E., Piktus, A., Petroni, F., Karpukhin, V., Goyal, N.,
  K\"{u}ttler, H., Lewis, M., Yih, W.t., Rockt\"{a}schel, T., Riedel, S.,
  Kiela, D.: Retrieval-augmented generation for knowledge-intensive nlp tasks.
  In: Larochelle, H., Ranzato, M., Hadsell, R., Balcan, M., Lin, H. (eds.)
  Advances in Neural Information Processing Systems. vol.~33, pp. 9459--9474.
  Curran Associates, Inc. (2020),
  \url{https://proceedings.neurips.cc/paper/2020/file/6b493230205f780e1bc26945df7481e5-Paper.pdf}

\bibitem{Lin2021pyserini}
Lin, J., Ma, X., Lin, S.C., Yang, J.H., Pradeep, R., Nogueira, R.: Pyserini: A
  python toolkit for reproducible information retrieval research with sparse
  and dense representations. In: Proceedings of the 44th International ACM
  SIGIR Conference on Research and Development in Information Retrieval. p.
  2356–2362. SIGIR '21, Association for Computing Machinery, New York, NY,
  USA (2021). \doi{10.1145/3404835.3463238},
  \url{https://doi.org/10.1145/3404835.3463238}

\bibitem{Liu2021inContext}
Liu, J., Shen, D., Zhang, Y., Dolan, B., Carin, L., Chen, W.: What makes good
  in-context examples for gpt-$3$? (2021). \doi{10.48550/ARXIV.2101.06804},
  \url{https://arxiv.org/abs/2101.06804}

\bibitem{webgpt}
Nakano, R., Hilton, J., Balaji, S., Wu, J., Ouyang, L., Kim, C., Hesse, C.,
  Jain, S., Kosaraju, V., Saunders, W., Jiang, X., Cobbe, K., Eloundou, T.,
  Krueger, G., Button, K., Knight, M., Chess, B., Schulman, J.: Webgpt:
  Browser-assisted question-answering with human feedback (2021).
  \doi{10.48550/ARXIV.2112.09332}, \url{https://arxiv.org/abs/2112.09332}

\bibitem{nogueira2020document}
Nogueira, R., Jiang, Z., Pradeep, R., Lin, J.: Document ranking with a
  pretrained sequence-to-sequence model. In: Findings of the Association for
  Computational Linguistics: EMNLP 2020. pp. 708--718. Association for
  Computational Linguistics, Online (Nov 2020).
  \doi{10.18653/v1/2020.findings-emnlp.63},
  \url{https://aclanthology.org/2020.findings-emnlp.63}

\bibitem{perez2020unsupervised}
Perez, E., Lewis, P., Yih, W.t., Cho, K., Kiela, D.: Unsupervised question
  decomposition for question answering. In: Proceedings of the 2020 Conference
  on Empirical Methods in Natural Language Processing (EMNLP). pp. 8864--8880
  (2020)

\bibitem{press2022measuring}
Press, O., Zhang, M., Min, S., Schmidt, L., Smith, N.A., Lewis, M.: Measuring
  and narrowing the compositionality gap in language models. arXiv preprint
  arXiv:2210.03350  (2022)

\bibitem{raffel2020exploring}
Raffel, C., Shazeer, N., Roberts, A., Lee, K., Narang, S., Matena, M., Zhou,
  Y., Li, W., Liu, P.J., et~al.: Exploring the limits of transfer learning with
  a unified text-to-text transformer. J. Mach. Learn. Res.  \textbf{21}(140),
  1--67 (2020)

\bibitem{reimers2019sbert}
Reimers, N., Gurevych, I.: Sentence-bert: Sentence embeddings using siamese
  bert-networks (2019). \doi{10.48550/ARXIV.1908.10084},
  \url{https://arxiv.org/abs/1908.10084}

\bibitem{Robertson1994OkapiAT}
Robertson, S.E., Walker, S., Jones, S., Hancock-Beaulieu, M., Gatford, M.:
  Okapi at trec-3. In: TREC (1994)

\bibitem{sachan2022questions}
Sachan, D.S., Lewis, M., Yogatama, D., Zettlemoyer, L., Pineau, J., Zaheer, M.:
  Questions are all you need to train a dense passage retriever. arXiv preprint
  arXiv:2206.10658  (2022)

\bibitem{ray2022mod}
Tay, Y., Dehghani, M., Tran, V.Q., Garcia, X., Bahri, D., Schuster, T., Zheng,
  H.S., Houlsby, N., Metzler, D.: Unifying language learning paradigms (2022).
  \doi{10.48550/ARXIV.2205.05131}, \url{https://arxiv.org/abs/2205.05131}

\bibitem{trivedi2022teching}
Trivedi, H., Balasubramanian, N., Khot, T., Sabharwal, A.: Teaching broad
  reasoning skills via decomposition-guided contexts (2022).
  \doi{10.48550/ARXIV.2205.12496}, \url{https://arxiv.org/abs/2205.12496}

\bibitem{wang2022}
Wang, X., Wei, J., Schuurmans, D., Le, Q., Chi, E., Narang, S., Chowdhery, A.,
  Zhou, D.: Self-consistency improves chain of thought reasoning in language
  models (2022). \doi{10.48550/ARXIV.2203.11171},
  \url{https://arxiv.org/abs/2203.11171}

\bibitem{wei2022CoT}
Wei, J., Wang, X., Schuurmans, D., Bosma, M., Ichter, B., Xia, F., Chi, E., Le,
  Q., Zhou, D.: Chain of thought prompting elicits reasoning in large language
  models (2022). \doi{10.48550/ARXIV.2201.11903},
  \url{https://arxiv.org/abs/2201.11903}

\bibitem{xiong2022longt5}
Xiong, W., Gupta, A., Toshniwal, S., Mehdad, Y., Yih, W.t.: Adapting pretrained
  text-to-text models for long text sequences (2022).
  \doi{10.48550/ARXIV.2209.10052}, \url{https://arxiv.org/abs/2209.10052}

\bibitem{xiong2020answering}
Xiong, W., Li, X.L., Iyer, S., Du, J., Lewis, P., Wang, W.Y., Mehdad, Y., Yih,
  W.t., Riedel, S., Kiela, D., Oğuz, B.: Answering complex open-domain
  questions with multi-hop dense retrieval (2020).
  \doi{10.48550/ARXIV.2009.12756}, \url{https://arxiv.org/abs/2009.12756}

\bibitem{Xu2016}
Xu, W., Napoles, C., Pavlick, E., Chen, Q., Callison-Burch, C.: {Optimizing
  Statistical Machine Translation for Text Simplification}. Transactions of the
  Association for Computational Linguistics  \textbf{4},  401--415 (07 2016).
  \doi{10.1162/tacl\_a\_00107}, \url{https://doi.org/10.1162/tacl\_a\_00107}

\bibitem{yang2018hotpotqa}
Yang, Z., Qi, P., Zhang, S., Bengio, Y., Cohen, W., Salakhutdinov, R., Manning,
  C.D.: Hotpotqa: A dataset for diverse, explainable multi-hop question
  answering. In: Proceedings of the 2018 Conference on Empirical Methods in
  Natural Language Processing. pp. 2369--2380 (2018)

\bibitem{zhu2021rr}
Zhu, F., Lei, W., Wang, C., Zheng, J., Poria, S., Chua, T.S.: Retrieving and
  reading: A comprehensive survey on open-domain question answering (2021).
  \doi{10.48550/ARXIV.2101.00774}, \url{https://arxiv.org/abs/2101.00774}

\end{thebibliography}

\end{document}